# A Tendon-Driven Wrist Abduction-Adduction Joint Improves Performance of a 5 DoF Upper Limb Exoskeleton - Implementation and Experimental Evaluation

Juwairiya S. Khan[1,6], Mostafa Mohammadi[1], Alexander L. Ammitzbøll[2], Ellen-Merete Hagen[2,4], Jakob Blicher[3] Izabella Obál[3], Ana S. S. Cardoso[1], Oguzhan Kirtas[5], Rasmus L. Kæseler[1] John Rasmussen[6], and Lotte N.S. Andreasen Struijk[1]

*Abstract- Objective:* Wrist function is essential in performing activities of daily living (ADLs). However, there is limited experimental evidence on the functional impact of wrist Abduction-Adduction (Ab-Ad) joint assistance in upper limb exoskeletons (ULEs) for rehabilitation. This study evaluates the effect of implementing an active wrist Ab-Ad joint in a five degree of freedom (DoF) ULE, EXOTIC[2] exoskeleton, to support individuals with severe motor impairments. *Methods:* A compact, lightweight wrist module with tendon-driven abduction and spring-driven adduction was integrated into the EXOTIC exoskeleton. Eight adults with no motor disabilities completed drinking and scratching tasks under randomized wrist-enabled and wrist-locked conditions along with a preliminary feasibility test in one individual with Amyotrophic lateral sclerosis (ALS). Kinematic and task performance metrics including wrist range of motion, task completion time, spillage and leveling metrics were assessed. *Results:* Implementing the wrist Ab-Ad DoF improved task success metrics. Spill incidence during the drinking task decreased from 56% to 3%, and leveling success for scratching task improved from 28% to 75%. *Conclusion:* Integrating wrist Ab-Ad assistance improved key functional task outcomes without increasing execution time. *Significance:* The study provides the experimental evidence that active wrist Ab-Ad control enhances task-level performance in exoskeleton-assisted ADLs.

*Index Terms—Upper-Limb Exoskeleton, Rehabilitation, Wrist Abduction-Adduction, Range of motion, Experimental Evaluation*

## I. INTRODUCTION

Spinal cord injury (SCI) affects an estimated 259,374 to 393,913 individuals in the United States, with approximately 18,421 new cases reported annually [1]. Cervical injuries account for more than half of all cases and often lead to incomplete or complete tetraplegia, which severely compromises independence, social participation, and performance of activities of daily living (ADL) [1], [2]. Although advances in surgical reconstruction, nerve transfer procedures, and regenerative interventions have improved upper limb outcomes for some individuals, many remain with persistent motor deficits that limit functional wrist and hand use [3], [4].

Assistive robotic devices and exoskeletons have been introduced as practical tools to support individuals with upper limb impairments due to SCI, stroke, or neurodegenerative disorders such as amyotrophic lateral sclerosis (ALS) [5]. The wrist is particularly critical for functional interaction with the environment because it orients the hand for grasping, stabilizing, and manipulating objects [6]. Limitations in wrist flexion-extension (FE), and abduction-adduction (Ab-Ad) can substantially restrict these tasks [7] and increase reliance on caregivers. Wrist exoskeletons aim to augment or restore these movements by providing actuation, gravity compensation, or controlled assistance for both daily use and therapeutic training [8], [9].

Designing effective wrist exoskeletons requires understanding the natural range of motion (ROM) and its variability across individuals [10]. Prior studies have characterized wrist kinematics in unimpaired populations, but fewer have examined ROM in individuals with SCI or other conditions where joint stiffness, muscle weakness, and compensatory strategies differ widely [11]. Moreover, movements such as wrist Ab-Ad, which are important for functional grasp orientation, are often underrepresented in both ROM analyses and device design [12], [13]. Incorporating user

This work has been submitted to the IEEE Transactions on Biomedical Engineering for possible publication. Copyright may be transferred without notice, after which this version may no longer be accessible.

[1]*Research supported by Aage og Johanne Louis-Hansens Fond.

[1]Juwairiya Siraj Khan (corresponding author), Mostafa Mohammadi, Ana S. S. Cardoso, Rasmus L. Kæseler and Lotte N.S. Andreasen. Struijk are with the Neurorehabilitation Robotics and Engineering research group, Center for Rehabilitation Robotics, Department of Health Science and Technology, Aalborg University, Gistrup, Denmark. (jsikh@hst.aau.dk , mostafa@hst.aau.dk, assc@hst.aau.dk , rlk@hst.aau.dk, naja@hst.aau.dk ).

[2]ALA (alexnl@rm.dk) and EMH (elhage@rm.dk) are with the Spinal Cord Injury Centre of Western Denmark, Regional Hospital of Viborg Denmark

[3] JB (j.blicher@rn.dk) and IO (i.obal@rn.dk) are with Aalborg University Hospital, Department of Neurology, 9000 Aalborg, Denmark

[4]EMH (elhage@rm.dk) is also with the Department of Clinical Medicine, Aarhus University, Denmark and Department of Brain Repair and Rehabilitation, UCL Queen Square Institute of Neurology, University College London, UK

5OK (okr@create.aau.dk) is with the Department of Architecture, Design and Media Technology, Aalborg University, Aalborg 9000, Denmark.

[6]Juwairiya Siraj Khan and John Rasmussen are with the Department of Materials and Production, Aalborg University, Aalborg East, Denmark (e-mail: jsikh@mp.aau.dk , jr@mp.aau.dk)

specific ROM data and clinician feedback is essential for ensuring comfort, safety, and functional benefit in assistive wrist robotics [13], [14]. In a recent study, we quantified wrist Ab-Ad ROM across participants with no motor disability (NMD), SCI, and ALS cohorts. Results revealed substantial inter-group and inter-individual variability, with differences exceeding 15-20° in some cases [10]. These findings highlighted that wrist deviation is not only essential for natural motion but also highly user-specific, underscoring the need for adjustable exoskeleton designs.

Despite these motivations, the field lacks controlled experimental evaluations demonstrating how including wrist Ab-Ad assistance affects functional performance in wearable ULEs during activities of daily living (ADLs). Most prior wrist exoskeletons have been actuated systems designed for rehabilitation, often bulky, and typically decoupled from larger multi-DoF arm systems [15]. Existing studies primarily focus on mechanism design, isolated joint evaluation, or ROM characterization, without assessing task-level impact in integrated multi-DoF systems [15]. The primary problem addressed in this work is therefore whether adding an active wrist Ab-Ad degree of freedom to a wearable arm exoskeleton measurably improves functional task performance and object manipulation during ADLs.

This is done in three steps. Firstly, by implementing a 6-DoF upper-limb exoskeleton by integrating a lightweight tendon-driven wrist Ab-Ad module [16] into the 5-Dof EXOTIC exoskeleton [17]. The 5-DoF EXOTIC exoskeleton previously enabled individuals with complete tetraplegia to perform highly prioritized ADLs. Secondly, we experimentally evaluate the effect of the added Ab-Ad module on the functional performance with 8 participants with NMD. We compare task execution with and without this joint during representative ADLs and analyze improvements in task success, wrist kinematics, and overall control quality. Thirdly, we perform a qualitative feasibility assessment with one individual with ALS. In contrast to our prior work on wrist Ab-Ad mechanism design [16] and population-specific ROM characterization [10], this study presents the first integrated system-level evaluation of wrist Ab-Ad assistance within a multi-DoF upper-limb exoskeleton during functional tasks.

The contributions of this study are as follows. First, we present the engineering integration and modeling of a lightweight tendon driven Ab-Ad joint within a multi DoF assistive arm exoskeleton. Second, we present a controlled human-subject evaluation of the 6-DoF EXOTIC$^2$ exoskeleton demonstrating the effect of wrist Ab-Ad assistance in a combined manual control and automatic leveling strategy enabling task-oriented wrist stabilization. We showcase this impact on functional performance metrics, including spillage reduction and object leveling success. We further present a preliminary qualitative proof-of-concept with one participant with ALS. Third, we link these findings to prior evidence of inter individual ROM variability and discuss how adaptive ROM design can better support diverse user populations.

## II. SYSTEM OVERVIEW.

The exoskeleton system used in this study is a six DoF exoskeleton (EXOTIC$^2$), as shown in Fig. 1., combining the four plus one DoF architecture of the EXOTIC platform (Upper arm and forearm + hand) with a newly developed tendon-driven wrist Ab-Ad exoskeleton joint. The joint structure consists of two shoulder DoFs: flexion-extension (FE) and internal external (IE) rotation, one elbow FE, one forearm rotation, and one wrist Ab-Ad DoF. A three-finger commercial tendon-driven active glove (CarbonHand, Bioservo, Sweden) is adapted to supply grasp assistance (sixth DoF), enabling participants to perform pick-and-place tasks without requiring intrinsic finger strength. Hand opening was enabled by means of passive elastic straps (Fig 1 b) added to the CarbonHand. Only the hand and wrist were physically coupled to the device using adjustable straps, while the upper and lower arm were left unconstrained, carried by braces of the exoskeleton. This configuration simplifies donning and reduces joint alignment errors at the anatomical joints during prolonged use. The EXOTIC$^2$ exoskeleton weighed approximately 3.9 kg. Both EXOTIC and EXOTIC$^2$ are tongue-controlled exoskeletons [18] but we only recruited NMD participants for this experimental evaluation study and therefore the exoskeleton ´was controlled using a handheld gamepad (Thrustmaster Dual Analog 5) that maps commands to joint velocities, with real time visual feedback and data logging. The detailed system description is presented in Fig. 2 (a). Additionally, a system feasibility assessment was done with one participant with ALS using a non-invasive tongue control interface, to provide intuitive and hands-free control of the exoskeleton, leveraging the preserved motor function of the tongue in individuals with severe motor impairments [19].

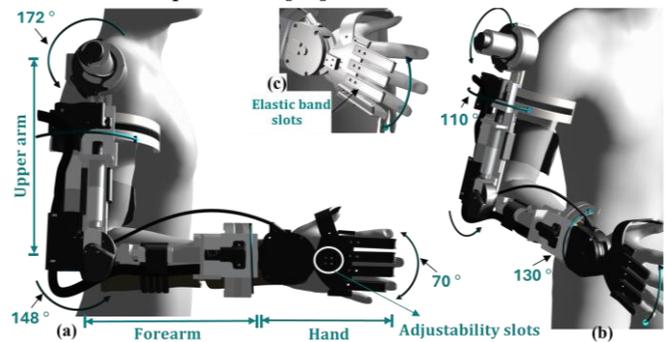

Fig. 1. CAD model of EXOTIC$^2$ exoskeleton and ROM of 5 DoFs. (a) side view with shoulder FE, elbow FE and wrist Ab-Ad and indications of adjustable links (upper arm, forearm and hand), (b) a view with shoulder IE rotation and forearm supination/pronation. (c) Wrist joint with finger slots for passive hand closing.

### A. Hardware

The EXOTIC exoskeleton is a 5-DoF wearable arm system developed to support shoulder and elbow movements during daily tasks. The shoulder mechanism is based on an anatomically inspired spherical joint representation, where glenohumeral FE and IE rotation are actively actuated while Ab-Ad is passively adjustable to maintain comfort and avoid collisions with the wheelchair environment. Actuation is achieved using a combination of Maxon brushless DC motors, harmonic drives, C-ring mechanisms, and spur gears, ensuring high torque density and low backlash across the workspace. The elbow joint provides powered FE driven through a worm-gear assembly, with adjustable upper-arm link length for anatomical alignment and user comfort. Forearm support includes a wrist-rotation C-ring module driven by a compact motor-gear unit. Hand opening and closing are achieved by means of a tendon-based soft exoskeleton glove (CarbonHand). These hardware elements were previously described in detail in [17].

To extend the EXOTIC exoskeleton with wrist deviation, we integrated a lightweight tendon-driven module originally developed in our prior work [16]. The design uses a hybrid actuation strategy wherein a Bowden cable provides active abduction torque, while a pre-loaded clock spring supplies passive adduction and maintains cable tension. The hand interface and base module are fabricated from polylactic acid (PLA) and PLA carbon fiber (PLA-CF) composites, with the rotational axis aligned to the wrist's anatomical Ab-Ad axis. A compact needle-bearing assembly minimizes friction. In addition to the encoders present in EXOTIC, an absolute miniature rotary magnetic encoder (RM08, RLS) was mounted at the wrist Ab-Ad joint. that provides real-time joint angle measurement.

The wrist module was dimensioned based on biomechanical requirements of ADLs, with target range of motion and assistive torque selected from literature [20], [21] and modeling to ensure functional support and user comfort [16]. Spring stiffness and pretension were chosen to provide continuous return torque, stable neutral positioning, and reduced motor effort during abduction. The cable transmission was designed to balance torque efficiency and friction losses while maintaining consistent tension across the range of motion, enabling smooth and reliable operation and adaptability to different users through adjustable interfaces. Further implementation details of the tendon-driven wrist Ab-Ad joint were provided in a previous study [16]. The physical space constraints in the EXOTIC exoskeleton such as the need for a compact and lightweight wrist joint that can be integrated in between the forearm and hand brace, without compromising the ergonomics and functionality, were also considered in the design and integration of the wrist Ab-Ad joint module in implementing a compact EXOTIC$^2$ exoskeleton. Since the idea was to make the EXOTIC$^2$ exoskeleton compact and lightweight, and cable-driven exoskeleton joints allow the placement of bulky actuators away from the joint, such as under the wheelchair of the user. Therefore, the choice of tendon-driven mechanism was majorly due to the space constraints of EXOTIC exoskeleton, overall weight of the system and its compactness.

During abduction, the motor winds the tendon, rotating the clock-spring housing and storing elastic energy. When the motor disengages or reverses, the clock spring provides a smooth restoring torque, returning the wrist to neutral or adducted posture. Because the spring is pretensioned, slack and backlash are effectively eliminated, ensuring smooth bidirectional control through a single active actuator. The actuation unit, including the motor and driver, is located proximally in a wheeled electronics box to reduce distal mass. This design yields 30° of available abduction and 40° of adduction, consistent with clinical ROM expectations and our population measurements [10], [22].

*B. Kinematic and Dynamic Modeling*

A simplified upper-limb model was developed in the AnyBody Modeling System [23] to estimate the wrist Ab-Ad torque requirements and transmission losses associated with Bowden-cable routing. The model comprised a spherical shoulder joint, elbow hinge, and an exoskeleton-driven wrist

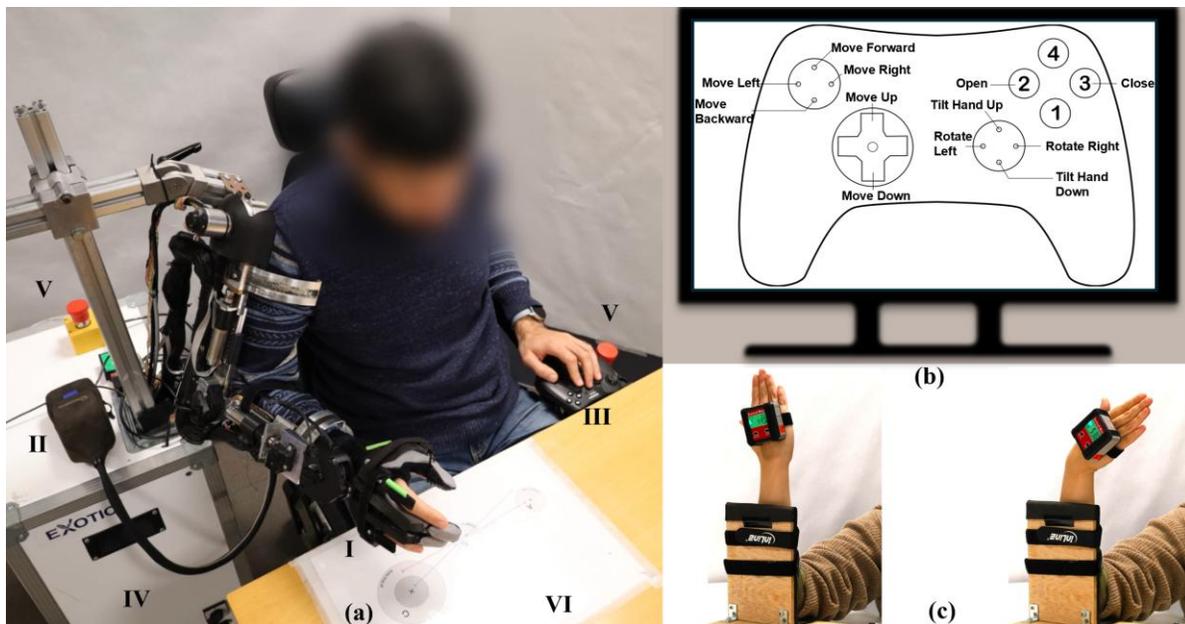

Fig. 2. (a) System overview during experimental evaluation. I: 6-DoF EXOTIC2 exoskeleton II: Exoskeleton Glove control unit, III: Gamepad control for NMD participants, IV: Home-made box containing all electronics boards and Motor box for wrist joint, V: Emergency stops for safety, VI: Table with experimental task positioning labels (b): Display with gamepad control commands (c) Wrist ROM measurement setup using a goniometer On Left side: Wrist abduction On Right side: Wrist adduction

deviation joint aligned anatomically to the user. The drinking trajectory was prescribed using a piecewise-linear hand path, allowing computation of the wrist deviation angle required to maintain cup horizontal alignment during elevation toward the mouth.

The external torque requirement at the wrist was estimated using a standard rotational dynamics model as in (1):

$$\tau_{req} = I\ddot{\theta} + b\dot{\theta} + \tau_{grav}, \quad (1)$$

where $I$ is wrist inertia, $b$ is passive damping, and $\tau_{grav}$ is the gravitational moment from the hand-cup system. Cable transmission losses were approximated using the capstan relation as in (2):

$$T_{out} = T_{in}\, e^{-\mu\theta_w}, \quad (2)$$

where $\mu$ is the Bowden friction coefficient and $\theta_w$ is the wrap angle along the tendon path. These estimates predicted approximately 8-12% tension loss under typical routing, which was incorporated into the required assistive torque and the required spring pretension.

The resulting torque envelope (≈12 N·mm/deg) guided the selection of the clock-spring stiffness for passive adduction return, which was selected to be 12.32 N·mm/deg. This kinematic and dynamic modelling approach further guided the engineering integration of the wrist Ab-Ad joint into the EXOTIC platform and informed the cable routing, and motor sizing used in the final prototype [16].

### C. Control and Software Architecture

The wrist joint operates in position control via a PID loop ($Kp = 100$, $Ki = 0.01$, $Kd = 10$), while shoulder, elbow, and forearm joints use velocity control. Identical gains were used across joints for implementation simplicity, rather than joint-specific optimization. Gamepad input allows intuitive, simultaneous multi-DoF movement, and all encoder data, command signals, and task timings are logged at up to 100 Hz. Controller gains were initially obtained using the EPOS Studio auto-tuning procedure and subsequently refined empirically to ensure stable tracking and smooth motion without oscillations. This approach is consistent with standard tuning practices [17] for EPOS-based systems. Similar control approaches have been used in prior work on the EXOTIC exoskeleton platform with communication over CAN and USB protocols [17].

During the experiment, the participants controlled the system through a gamepad interface. Horizontal-plane motion of the hand was regulated via a two-axis joystick providing continuous velocity commands, while a separate lever enabled vertical translation along the z-axis. A dual-axis paddle-controlled wrist rotation (pronation-supination) and wrist deviation (Ab-Ad). Two additional buttons were dedicated to grasping and releasing the object via the glove. The detailed gamepad layout is presented in Fig. 2 (b) For safety and user comfort, maximum hand linear velocity was limited to 4 cm/s, and wrist angular velocity to 0.2 rad/s based on pilot tests to ensure safe and smooth motion [17], [18].

When the wrist Ab-Ad joint was enabled, in addition to the user-driven manual control, an automatic leveling controller maintained the hand frame's orientation in the global reference frame while the hand was moving during task execution. The controller continuously monitored the end-effector orientation and applied corrective wrist deviation commands to counteract unwanted tilt, particularly important in the drinking task for keeping the cup level. A simple proportional law was used as in (3),

$$\tau_w = K_p(\theta_{ref} - \theta_{wrist}) \quad (3)$$

where $\theta_{ref} = 0°$ denotes the desired horizontal alignment. $\theta_{ref}$ is 0 by default (when the system starts), which roughly corresponds to being parallel with the plane of the table. The user controls the wrist, by specifying $\theta_{ref}$ - thereby ensuring that any adjustments made by the user is accounted for in the automatic levelling. The controller operated in parallel with the user's joystick input, providing gentle stabilization without restricting voluntary movement. This feature ensured stable object orientation during reaching and transport phases and was part of the wrist-enabled condition evaluated experimentally.

### D. Safety and Calibration

Mechanical stops adjusted to the participant's individual wrist Ab-Ad ROM and software limits are implemented to avoid over-stretching the joint. Furthermore, current limits control the motor torque. Emergency-stops (as shown in Fig. 2 (a)) provided both for the participant and the experimenter ensure safe operation. Each session begins with a joint calibration routine ensuring alignment of the exoskeleton and the user's wrist neutral axis.

## III. EXPERIMENTAL METHODS

### A. Participants

Eight NMD adults (7 male, 1 female; mean age $28.75 \pm 5.73$) participated in the experimental study. One participant was left-handed. Participants had no history of musculoskeletal or neurological disorders affecting arm/wrist function. Additionally, a preliminary feasibility test was conducted with one individual with ALS to assess the concept. The participant was a male (age 73 years) diagnosed 1-year and 10 months prior, with an ALSFRSr score of 43. All provided written informed consent, and procedures were approved by the regional ethics committee (The Scientific Ethics Committee for the North Jutland region, VEK, Protocol number: 20220029 and 20230044 for NMD and ALS participant respectively).

### B. Experimental Protocol

A within-subject design was used, with each participant performing two experimental conditions: 1) Enabled condition: wrist Ab-Ad active (both manual control and automatic-leveling were active), free motion. 2) Disabled condition: wrist mechanically locked in neutral state (by means of software which disabled the motor driver). The other exoskeleton joints and the experimental setup remained consistent during both experimental conditions. The system overview during experimental evaluation is shown is Fig. 2. Condition orders were randomized and participants completed familiarization trials before data collection. This study included one experimental session and lasted for about an hour.

In addition to the main protocol, a preliminary feasibility observation was conducted with one individual with ALS. Due to the participant's motor impairment, the system was controlled using a non-invasive tongue control interface [19] rather than a gamepad. The participant performed ADLs such as retrieving a fallen bottle and several tasks of the ARAT test (including grasp test: 5cm block, grip test: tumbler, 2.5 cm wide tube and gross movements: move hand to mouth) [24] with wrist assistance and provided qualitative feedback through a short questionnaire. This observation was exploratory and not included in the statistical analysis. Active wrist Ab-Ad ROM were measured for each NMD participant during the prior information meeting, while passive ROM was measured for ALS participant. A digital goniometer was used for joint motion measurements shown in Fig 2. (c), and their ROM was used to adjust the joint limits in EXOTIC[2] [10].

### C. Functional Tasks

Two functional tasks, drinking and scratching were carried out in this experimental study. These tasks were selected based on prior user-centered studies and desires from end-users obtained at interviews and User-Panel meetings with users with SCI and ALS at Center for rehabilitation Robotics, CREROB, at Aalborg University [13], [25]. The detailed experimental protocol is shown in Fig. 3 with tasks, conditions, trials and outcomes.

#### 1. Drinking Task

Participants reached for a cup placed inside a secondary collection container (bottle) with calibration markings (Fig 3 e), to avoid spilling of liquid on the participants and the exoskeleton setup. The participants grasped the bottle, lifted it, and drank through a straw as shown in Fig. 3. Spillage was classified via post-trial photographic inspection of the collection container. If evidence of liquid residue or visible drop pattern was present in the container photograph (Fig 3 e), the trial was scored as "spill"; otherwise it was scored "no-spill."

#### 2. Scratching and Leveling Task

Participants reached to pick up a cylindrical scratch stick, touched their nose with tip of the stick, and then placed the stick onto a flat circular marker requiring leveling of stick base with the table (position B) as shown in Fig. 3. Final outcomes were categorized as: (a) exoskeleton-only success: when participants succeeded in leveling the stick only with the exoskeleton (Fig 3 e), (b) human-assisted success: when participants succeeded in leveling but with the support of their hand or with help of their own hand grasp force, (c) not leveled: when stick could not be levelled and it ultimately fell on table, (d) grasp failure: stick not leveled due to poor grasp leading to drop. Each task was recorded with video and participants were asked for each trial if they assisted in addition to the exoskeleton's assistance. The observations were saved in a case report form for processing.

### D. Data Collection

The following parameters were assessed during the study:

1) Joint angles, task completion time (TCT), and trial outcomes were logged. Wrist deviation was sampled continuously, enabling calculation of RMS, directional peaks, and ROM.

2) *Peak ROM Extraction*

For every enabled trial, the following directional metrics were extracted from raw Ab-Ad trajectories: (a) Abduction peak (°): maximal positive deviation, (b) Adduction peak (°): minimal negative deviation, (c) Total ROM (°): $\theta_{\max} - \theta_{\min}$. Peak values were computed per trial and aggregated at the participant level using the maximum across trial.

3) *Pre-ROM Measurement*

A separate baseline active wrist ROM (abduction and adduction) for each participant was measured as shown in Fig. 2 (c). During ROM measurements, the wrist neutral was subjective based on each participant's perception of their neutral wrist. These values were used to compare natural ROM to the ROM expressed during task execution while assisted by the device.

4) *Grasp and Release Deviation*

For each trial, wrist deviation was sampled at: Grasp (start of manipulation) and Release (end of scratching task only). The resulting distributions were visualized using violin plots with superimposed scatter, separately for drinking (grasp only) and scratching (grasp and release). This allowed evaluation of whether the device constrained or biased wrist orientation during contact transitions.

### E. Data Processing and Statistics

Normality was assessed with the Shapiro-Wilk test; paired-t tests (α = 0.05) were used when residuals were approximately normal, otherwise Wilcoxon signed-rank tests were used. Reported p-values are two-tailed. Effect size was estimated as Cohen's d for paired samples. For categorical outcome comparisons (e.g., leveling task outcomes), a chi-square test of independence was applied to assess differences in outcome distributions between conditions. The sample size was selected based on feasibility and the expectation of large effect sizes in a within-subject design. For paired comparisons, sample sizes on the order of 8-10 participants are generally sufficient to detect large effects (Cohen's d ≈ 0.8) at α = 0.05 with 80% power [26]. This is consistent with prior experimental studies in upper-limb exoskeleton evaluation, which typically employ small participant cohorts due to system complexity and testing constraints [15], [17]. All processing was performed in Python with standard numeric libraries. ROM, abduction and adduction peaks are computed as shown in (4, 5 and 6):

$$\text{ROM} = \max(d_i) - \min(d_i) \quad (4)$$
$$\text{Abduction peak} = \max(d_i) \quad (5)$$
$$\text{Adduction peak} = \min(d_i) \quad (6)$$

## IV. RESULTS

All eight participants completed the drinking and scratching tasks with and without the wrist Ab-Ad joint active. Detailed findings for task performance, kinematic analysis, and user observations are presented below..

### A. Task Performance Outcomes

#### 1. Drinking task spillage

Enabling the wrist DoF resulted in a substantial reduction of spillage. With the wrist DoF disabled, 18 out of 32 trials were recorded as 'spill' events, while only one 'spill' occurred when wrist was enabled. Across all participants, spill probability decreased from 56% (wrist disabled) to 3% (wrist enabled (p = 0.0005) indicating improved functional task performance through enhanced cup orientation control when the wrist was enabled (Fig. 4 a).

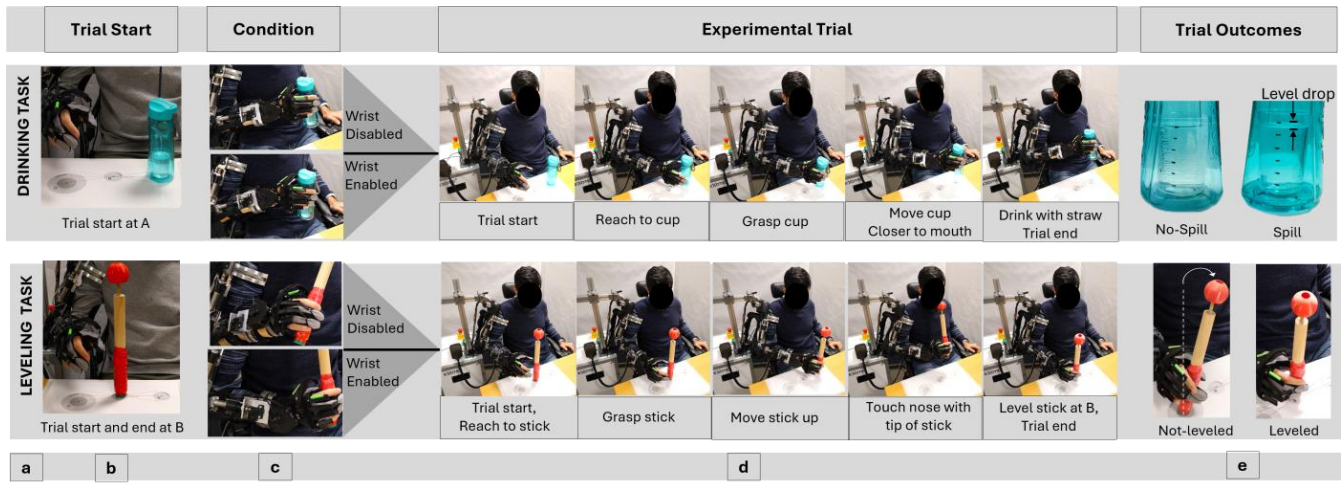

Fig. 3. Experimental Protocol for NMD participants indicating two tasks (a) drinking with a cup (placed in a container) and leveling scratching stick, (b) Trial start positions, (c) Conditions: With and without wrist Ab-Ad joint, (d) image series of the experimental trial performed by the participant and (e) the trial outcomes.

*2. Scratching and Leveling Task outcomes*

Leveling performance improved significantly when the wrist DoF was enabled. A chi-square test revealed a strong effect of active wrist ($\chi^2 = 27.71$, $p = 4.17 \times 10^{-6}$). Exoskeleton-only success increased substantially from 28% to 75%, with the wrist enabled while Not-leveled outcomes decreased. Participant-specific paired tests confirmed improvements in both Exoskeleton-only-success ($p = 0.0011$) and reduction in failure ($p = 0.0185$).

*3. Task completion time*

Task completion time did not differ significantly between conditions (paired t-test, drinking task: p = 0.267 and leveling task: p = 0.487), indicating that enabling the wrist did not impose additional cognitive or kinematic burden. Participants often reported the opposite: that motions felt more natural and required less compensatory alignment.

*B. Wrist Kinematic Metrics*

*1. Wrist Range of Motion (ROM)*

Across participants, wrist deviations during both tasks remained well within each individual's pre-ROM envelope Fig. 5 (c), A consistent task-dependent pattern emerged: scratching/leveling elicited greater abduction across nearly all participants, whereas drinking produced larger adduction excursions, though still inside the measured negative pre-ROM boundary (Fig 5 c). This differentiation reflects functional biomechanics: scratching movements naturally require radial deviation (abduction) specially to scratch the nose by titling the wrist, while drinking motions involve ulnar deviation (adduction) during bringing the cup close to mouth and orienting it. Although all participants completed the same tasks under the same protocol, the magnitude and distribution of deviations varied across individuals (Fig 5 c). Some participants operated with broader ROM bands, while others remained close to neutral. Importantly, this variability may very well reflect individual exoskeleton control strategies. Overall, the results show that the exoskeleton-maintained wrist motion within each participant's pre measured Wrist ROM while allowing for task-specific mechanics and individual control strategies. Furthermore, during the wrist disabled state the joint remained in the locked state and the mean joint angle across all participants remained 0°.

*2. Drinking Task - Grasping-time Wrist Deviation*

The drinking trials cup grasping-time wrist deviation violin plot in Fig. 5 (a) revealed distinct participant-specific wrist

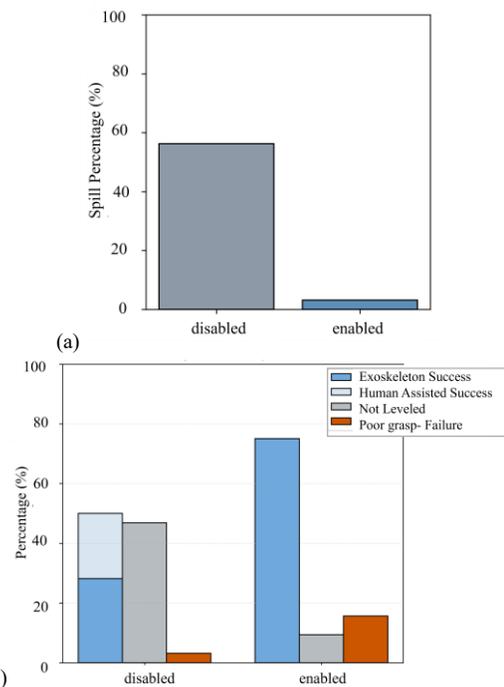

Fig. 4. Task outcomes during experimental trials of NMD participants (a) Spill outcomes during drinking task, (b) leveling outcomes during scratching task

strategies, with some participants consistently adopting small adduction (e.g., P1, P7, P8). Others showing broader, multi-modal distributions (e.g., P2-P4, P6), suggesting more variable grasp postures. Most deviations remained moderate (-30° to +20°), indicating that the exoskeleton did not force the wrist into extreme compensatory positions

*3. Scratching Task - Grasp/Release time wrist deviations*

For scratching trials, stick release time deviations of wrist were generally smaller than stick grasp deviations, implying that participants relaxed the wrist closer to neutral during task completion (Fig. 5 (b)). The grasp-release deviation distributions were asymmetric for several participants (e.g. P3-P6), which is expected because the scratching motion requires radial deviation during target contact and ulnar return during release. Violin widths demonstrated low variance for participants with consistent control strategies (P7-P8) and high variance for those adopting multiple path during task and control strategies (P2-P4).

### C. Subjective Responses

Participants consistently noted that wrist-enabled motion felt more natural, especially during the cup approach in the drinking task. One participant quoted, "Can I tilt my hand?" while wrist Ab-Ad was disabled during drinking task but the participant felt like exploiting the joint. While another participant addressed, "Wrist tilting helped before!". During the scratching and leveling task, most participants remarked that controlling the stick orientation was easier when wrist deviation was available.

Additionally, to provide initial insight into applicability in the target population, a preliminary feasibility observation was conducted with one individual with ALS. The participant was able to complete the tasks with wrist assistance and reported improved ease of object orientation and task execution. Qualitative questionnaire feedback from the user further indicated that wrist Ab-Ad assistance was perceived as important for performing ADLs, device weight and comfort were rated favorably, with no significant pressure points reported. Responses were consistently positive (average ≈ 1.5 on a 1-10 Likert scale, where lower values indicate more favorable ratings) as shown in Table I. This observation is qualitative and should be interpreted with caution because it is an exploratory preliminary assessment of the system.

TABLE I. Usability Questionnaire: Likert scale score from 1 to 10 by a user with ALS, during the preliminary assessment of the EXOTIC$^2$ exoskeleton

| Questions | Score: 1 | Score: 10 | Rating |
|---|---|---|---|
| How important was it to include the Ab-Ad joint for performing ADLs | Necessary | Not necessary | 1 |
| Did you feel the weight of the wrist exoskeleton too heavy on your hand? | Not heavy | Unacceptably heavy | 3 |
| Was it easy to don and doff the exoskeleton? | Easy | Unacceptably difficult | 2 |
| What do you think of the functionality of the wrist exoskeleton? | Useful | Not Useful | 2 |
| Are there any pressure points or discomfort associated with wearing the wrist exoskeleton for an extended period. | No discomfort | Unacceptable discomfort | 1 |

## V. DISCUSSION

This study provides the first systematic demonstration that adding wrist abduction-adduction to a wearable upper-limb exoskeleton significantly improves ADL performance and wrist kinematics. When the wrist DoF was disabled, the participants majorly rely on proximal joints to orient objects. Enabling the wrist, both manual and automatic leveling control methods, supported more natural task-specific kinematic patterns and clear task-specific directional control. Both manual control of the wrist Ab-Ad joint and automatic hand leveling appeared to support maintenance of object orientation during the wrist-enabled condition, consistent with coordinated multi-joint control strategies in human manipulation tasks [27]. As the wrist-enabled condition includes both the additional degree of freedom and an automatic leveling controller, their individual contributions were not fully separated in this study, but the leveling controller would not have been able to change the wrist Ab-Ad level if there had been no Ab-Ad joint implemented.

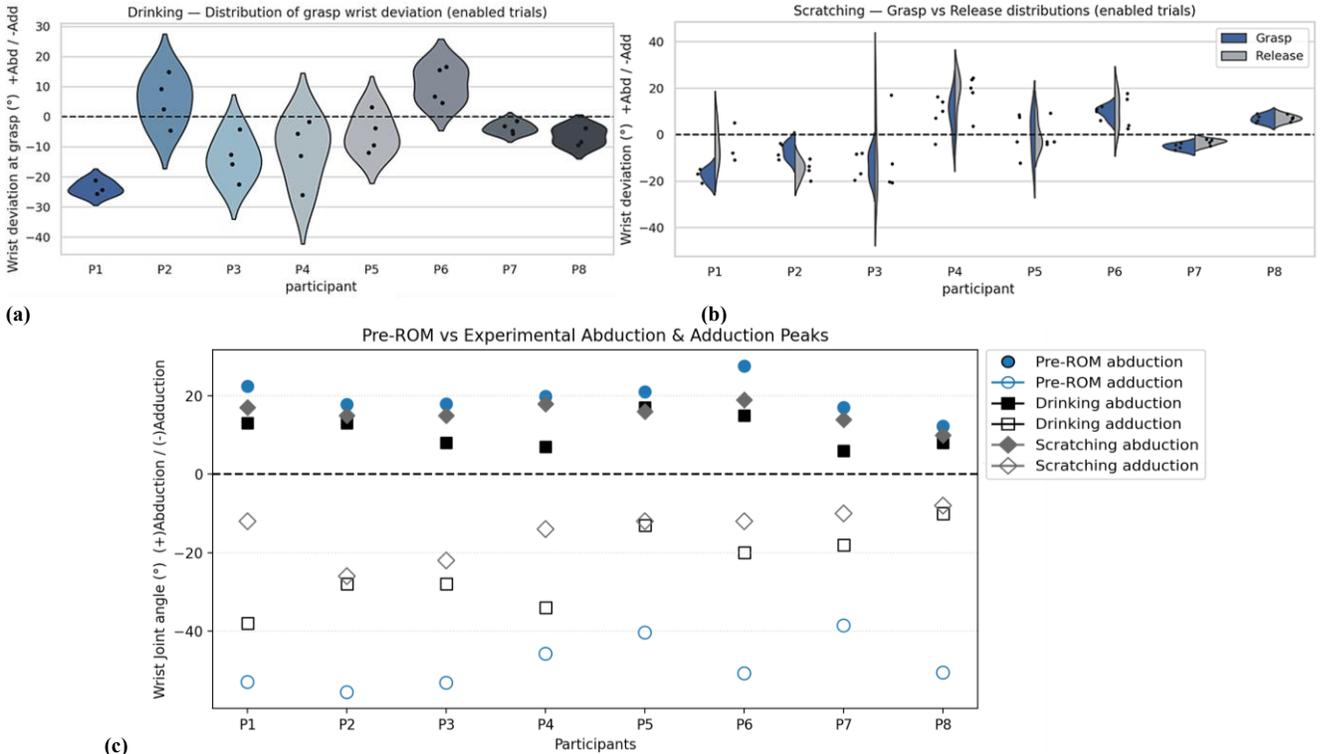

Fig. 5. ROM results for participants with NMD (a) Distribution of grasp for wrist Ab-Ad during drinking task (b) Distribution of grasp and release for wrist Ab-Ad during scratching task (c)Pre-measured ROM and experimental ROM during drinking and scratching tasks

Wrist-enabled trials exhibited substantially fewer spills (one vs 18 out of 32 trials) and significantly improved leveling outcomes. These results are biomechanically intuitive: cup orientation and object leveling tasks are fundamentally wrist-driven actions, and absence of the wrist DoF imposes compensatory strategies that are less stable and ergonomically suboptimal [7], [12]. Experimental wrist Ab-Ad peaks stayed well within each participant's pre-ROM limits, indicating that the exoskeleton supported stable mid-range postures rather than pushing users toward extreme ROMs.

Participants with larger wrist-strategy variability (e.g., P2-P4) showed more variation in leveling performance, suggesting that adaptive wrist deviation may support smoother functional interaction. Across both tasks, no signs of unsafe or compensatory motion were observed. Participants avoided extreme adduction, stayed far from extreme ROMs, and typically released objects near neutral-patterns.

The grasp-release distributions provide additional evidence that the exoskeleton supported stable but natural wrist behavior across tasks. During drinking, participants showed clear, individualized grasp strategies: some consistently adopted small adduction angles, while others displayed broader or multi-modal distributions, yet all remained within moderate deviation ranges. This suggests that the assistance did not enforce uniformity but helped participants operate comfortably within their preferred kinematic patterns without drifting into compensatory extremes. In scratching, the expected biomechanical asymmetry between grasp and release was preserved, participants typically contacted the surface with greater abduction and returned toward neutral during release. The narrower distributions observed in participants with consistent control strategies contrasted with the wider envelopes in those using multiple motor solutions, reflecting genuine inter-subject variability rather than device-induced inconsistency. Together, these findings indicate that the exoskeleton stabilized the wrist enough to support functional manipulation while still allowing for the user's natural grasp-release strategy.

Importantly, enabling the wrist did not increase task completion time, suggesting that the additional DoF does not impose cognitive overhead. Rather, participants reported improved naturalness and reduced effort. These results reinforce the need for personalized wrist ROM settings. In our prior study [10], we showed that Ab-Ad ROM varies substantially across NMD, SCI, and ALS cohorts. The current experimental findings extend that argument by demonstrating that when Ab-Ad assistance is available, users exploit the degree of freedom in task-specific ways, producing measurable improvements in object stability and task success. Together, the two studies provide complementary evidence: anatomical ROM varies substantially between individuals [10], and implementing wrist Ab-Ad assistance yields functional benefits regardless of a user's baseline ROM. The tendon-driven wrist module presented here is compatible with such personalization, and future work will incorporate adjustable mechanical limits and adaptive control strategies informed by user-specific kinematic profiles.

While the observed trends are promising, results should be interpreted with appropriate caution since this exploratory study was limited to a small healthy cohort for quantitative evaluation and analysis. Further, except from the study with the individual with ALS, the study used a manual gamepad control rather than intention-based interfaces, to ensure simple and consistent control. The observed improvements in object orientation and task success are primarily driven by the added wrist Ab-Ad DoF, which reduces reliance on compensatory proximal strategies. This is particularly relevant for individuals with SCI or ALS, who often lack active wrist control and depend on assistive devices to achieve stable hand orientation during ADLs. The selected tasks in the study were based on prior user-centered studies with individuals with SCI and ALS [13], [25], supporting their clinical relevance. However, there were occasional grasp failures due to exoskeleton glove limitations.

As a preliminary step toward clinical validation, the system was tested with one individual with ALS to assess feasibility of use. The participant used a tongue control interface [18] and was able to complete the tasks with wrist Ab-Ad joint assistance and reported that wrist motion contributed to improved object orientation and task execution. While this single-user observation is qualitative and not intended as quantitative evidence, it provides initial support for the applicability of proposed wrist Ab-Ad assistance in the target population.

Future work will integrate improved hand modules, adaptive ROM tuning, and intention-driven control. Furthermore, as gamepad control is not suitable for SCI or ALS participants, EXOTIC$^2$ exoskeleton with integrated wrist module will further be evaluated in clinical cohorts using alternative control modalities such as tongue control [18]to examine long-term usability, fatigue, and independence benefits and clinical relevance [5], [15], [17].

## VI. CONCLUSION

Integrating a lightweight tendon-driven wrist abduction-adduction mechanism into a wearable upper-limb exoskeleton significantly improved key functional task outcomes during activities of daily living. Across both drinking and leveling tasks, the 6-DoF EXOTIC$^2$ exoskeleton enabled substantial increases in wrist ROM and introduced task-dependent control that was absent when the wrist was mechanically locked. These kinematic improvements translated directly to functional benefits, including a clear reduction in spill events and a marked increase in successful scratch stick leveling, without increasing task completion time. The qualitative assessment from one participant with ALS also suggests the clinical importance of wrist Ab-Ad joint in ADLs.

The findings highlight the fundamental role of wrist deviation in achieving natural hand orientation and stable object manipulation and support the inclusion of wrist Ab-Ad assistance in future exoskeletons intended for individuals with SCI and ALS. Overall, this study establishes wrist deviation assistance as a critical functional component for next-generation upper-limb exoskeletons tailored to users with neurological impairment.

ACKNOWLEDGMENT

This work has been funded by the Aage and Johanne Louis-Hansens Foundation (Grant No. 20-2B- 7273), Denmark at Aalborg University. We thank Stefan Hein Bengtson and Irina Simona Socol-Lontis for help during experiments.